\documentclass[11pt]{article}

\usepackage[utf8]{inputenc}
\usepackage[T1]{fontenc}
\usepackage{mathptmx}            
\usepackage[margin=1in]{geometry}
\usepackage{amsmath,amssymb}
\usepackage{graphicx}
\usepackage{booktabs}
\usepackage{array}
\usepackage{caption}
\usepackage{subcaption}
\usepackage{xcolor}
\usepackage{tikz}
\usetikzlibrary{arrows.meta,positioning,shapes.geometric,calc,fit,backgrounds}
\usepackage{enumitem}
\usepackage{authblk}
\usepackage[hidelinks]{hyperref}
\usepackage{microtype}

\setlist{itemsep=2pt,topsep=3pt,parsep=0pt}

\definecolor{navy}{HTML}{1F3A5F}
\definecolor{rust}{HTML}{B5462A}
\definecolor{teal}{HTML}{2F7E7E}
\definecolor{lgrey}{HTML}{EDEDED}

\title{\vspace{-1.2cm}\bfseries Data Collection\\
for Training Quality-Control AI in Carpet Manufacturing:\\
A Design Proposal Grounded in a Six Sigma Project in Woven Carpet Production}

\author[1]{Akbar Erkinov}
\affil[1]{\small Independent Researcher}
\date{}

\begin{document}
\maketitle
\vspace{-0.8cm}

\begin{abstract}
\noindent
Visual inspection remains the dominant quality-control practice in woven and tufted
carpet production, yet it is slow, subjective, and inconsistent at the line speeds and
widths of modern looms. We present a design proposal for an in-line machine-vision system
whose primary purpose is twofold: to inspect the carpet web in real time, and---equally
important---to systematically \emph{collect and label} images of defect patterns so that
increasingly capable quality-control models can be trained over the life of the
installation. The proposal is grounded in a concrete industrial setting: a Six Sigma
(DMAIC) project at a woven-carpet production facility that anticipated a production bottleneck after the
installation of additional weaving machines, with a reported baseline defects-per-million
(DPMO) of roughly $2.0\times10^{4}$ and a financial exposure on the order of
$5\times10^{4}$~EUR per day. We describe an imaging subsystem based on synchronized
line-scan cameras with combined bright-field and grazing illumination, derive the
resolution and throughput budget required to resolve fine structural defects across a
multi-metre web, and define a carpet-specific defect taxonomy. We then lay out a staged
modelling strategy that begins with unsupervised anomaly detection trained on defect-free
material---the regime exemplified by the carpet category of the MVTec Anomaly Detection
benchmark---and matures, through a human-in-the-loop ``annotation flywheel,'' into
supervised detection and segmentation models. Finally, we connect detection performance to
the DMAIC objectives, showing how reductions in escaped defects map onto DPMO reduction and
an improved process sigma level. The contribution is an end-to-end, deployable blueprint
that treats data collection as a first-class engineering goal rather than an afterthought.
\end{abstract}


\section{Introduction}
Carpet is produced as a continuous web that can be several metres wide and is advanced
through weaving, finishing, and backing (gluing) stages at substantial linear speed. Quality
is judged largely on appearance: a single broken pick, a band of off-shade pile, a stain, or
a delaminated patch of backing can downgrade or reject a roll that represents hours of
machine time and considerable raw material. In most plants the responsibility for catching
these faults still rests on human inspectors, who view the moving web or sample cut pieces.
Manual inspection scales poorly: attention degrades over a shift, inspectors disagree with
one another, fine or low-contrast faults are missed, and only a fraction of the total surface
can be examined when the line runs fast. As throughput rises, the gap between what is produced
and what is actually inspected widens.

This gap was precisely the motivation for a Six Sigma project at a woven-carpet production facility. The project
charter identified a likely bottleneck arising from the installation of additional weaving
machines: woven output would increase while downstream capacity---including inspection---would
not. The charter recorded a baseline defects-per-million-opportunities (DPMO) of about
$2.0\times10^{4}$, a reported process sigma level of $2.33$, a target increase in output from
$7.0\times10^{5}$ to $1.0\times10^{6}$ square metres over the project horizon, and an estimated
exposure of roughly $50{,}000$~EUR per day if the bottleneck were left unaddressed. The
charter scoped the relevant process as running ``from weaving the carpet to gluing.''

In this paper we argue that camera-based, AI-assisted inspection is the natural lever for this
problem, and we develop the argument into a concrete engineering proposal. Crucially, we do
not treat the inspection model as something that exists before deployment. High-quality
industrial models require data that simply does not exist on day one: defects are rare,
diverse, and specific to a plant's yarns, dyes, looms, and patterns. The central idea of this
proposal is therefore to design the camera system and the surrounding workflow so that
\emph{data collection of defect patterns is a primary output}, feeding a virtuous cycle in
which each week of production improves the dataset and the dataset improves the models.

The remainder of the paper is organized as follows. Section~\ref{sec:background} reviews
carpet defects and the state of the art in automated fabric inspection. Section~\ref{sec:case}
states the problem within the DMAIC framework. Section~\ref{sec:imaging} specifies
the imaging and acquisition subsystem and derives its resolution and throughput budget.
Section~\ref{sec:taxonomy} defines a carpet defect taxonomy. Section~\ref{sec:data} describes
the dataset-construction methodology and the annotation flywheel. Section~\ref{sec:models}
presents the staged modelling strategy. Section~\ref{sec:sixsigma} ties detection performance
to DPMO and the sigma level. Section~\ref{sec:deploy} discusses deployment, and
Sections~\ref{sec:limits}--\ref{sec:conclusion} cover limitations and conclusions.

\section{Background and Related Work}
\label{sec:background}

\subsection{Carpet and the appearance of defects}
Woven carpet (for example Wilton and Axminster constructions) is formed by interlacing warp
and weft yarns with a raised pile, after which a secondary backing is bonded with adhesive.
Faults can originate at every stage. Weaving introduces structural faults such as broken or
missing picks, floats, double picks, and tension bands; dyeing and yarn variation produce
shade variation, streakiness, and foreign-fibre contamination; handling produces oil marks,
stains, and embedded debris; and the backing stage can produce delamination, adhesive
starvation, or bubbling. Many of these faults are subtle, are only visible under particular
lighting, or extend over large areas with low local contrast, which is exactly what makes
consistent manual detection difficult.

\subsection{Automated visual inspection of fabrics}
Automated fabric inspection has a long history. Early systems relied on hand-engineered
features---statistical texture descriptors, structural (primitive-and-placement) models,
spectral methods using Fourier or Gabor filtering, and model-based approaches---surveyed
comprehensively by Kumar~\cite{kumar2008} and by Ngan, Pang and Yung~\cite{ngan2011}. These
methods can work well for regular, repetitive textures but tend to be brittle: they require
careful per-product tuning and degrade on the patterned, multi-colour, high-pile surfaces
typical of carpet.

\subsection{Deep learning for defect detection}
The shift to learned representations followed the broader success of convolutional neural
networks in image recognition~\cite{krizhevsky2012,lecun2015}. Architectures such as
VGG~\cite{vgg}, ResNet~\cite{resnet}, and EfficientNet~\cite{efficientnet} provide strong
backbones for transfer learning; detectors such as Faster~R-CNN~\cite{fasterrcnn} and the
YOLO family~\cite{yolo} localize faults with bounding boxes; and encoder--decoder
segmentation networks such as U-Net~\cite{unet} and instance models such as
Mask~R-CNN~\cite{maskrcnn} produce pixel-precise fault maps. Recent surveys of fabric defect
detection document a clear migration from hand-crafted pipelines to deep models over the past
two decades~\cite{kahraman2023,li2021survey}, including early autoencoder
approaches~\cite{mei2018} and very recent real-time YOLO-based systems for textile inspection
on edge hardware~\cite{electronics2025}.

\subsection{Unsupervised and self-supervised anomaly detection}
A distinct and, for our purpose, decisive line of work targets the \emph{cold-start} problem:
learning to flag defects when only defect-free examples are available for training. The MVTec
Anomaly Detection (MVTec~AD) benchmark crystallized this setting and, notably, includes a
\emph{carpet} texture category alongside other industrial surfaces~\cite{mvtec2019}. Methods
such as PaDiM~\cite{padim} and PatchCore~\cite{patchcore} model the distribution of normal
image patches using features from pre-trained networks and score test patches by their
distance from that distribution; PatchCore reports image-level detection AUROC above $99\%$ on
MVTec~AD. Self-supervised approaches synthesize plausible defects on normal images---for
example CutPaste~\cite{cutpaste} and the reconstruction-based DRAEM~\cite{draem}---to train
discriminative localizers without real defect labels. These techniques are what allow a
freshly installed inspection line to be useful before a labelled defect dataset exists, and
they are central to our data-collection strategy.

\section{The Case Study and Problem Framing}
\label{sec:case}
The project followed the Define--Measure--Analyse--Improve--Control (DMAIC) structure of
Six Sigma~\cite{pyzdek}. We summarize the charter and then position the proposed system within
each phase (Table~\ref{tab:charter}).

\begin{table}[t]
\centering
\caption{Charter parameters of the case-study project and the role of camera-based AI within
the DMAIC phases.}
\label{tab:charter}
\small
\begin{tabular}{@{}p{0.30\linewidth} p{0.62\linewidth}@{}}
\toprule
\textbf{Charter element} & \textbf{Recorded value / role of vision system}\\
\midrule
Problem & Anticipated bottleneck after new weaving machines: output rises while downstream
(including inspection) capacity is unchanged.\\
Baseline DPMO & $\approx 2.0\times10^{4}$ defects per million opportunities.\\
Reported sigma level & $2.33$.\\
Throughput target & $7.0\times10^{5} \rightarrow 1.0\times10^{6}$ m$^2$ over the horizon.\\
Financial exposure & $\approx 50{,}000$~EUR/day if unaddressed.\\
Process scope & ``From weaving the carpet to gluing.''\\
\midrule
Define & Inspection becomes the controlled, instrumented checkpoint of the scoped process.\\
Measure & Cameras provide $100\%$ surface coverage and a continuous, objective defect record
in place of sampled manual checks.\\
Analyse & Localized, classified defect maps reveal which loom, shift, or yarn lot drives
specific fault families.\\
Improve & Real-time alerts shorten the detect-to-correct loop; faults are caught at weaving
rather than after gluing.\\
Control & Standardized thresholds, dashboards, and statistical process control sustain the
gain and prevent regression.\\
\bottomrule
\end{tabular}
\end{table}

The key observation is that inspection sits at the heart of the scoped ``weaving-to-gluing''
process. If output grows by more than $40\%$ (from $7.0\times10^{5}$ to $1.0\times10^{6}$~m$^2$)
without a corresponding increase in inspection capacity, escaped defects must rise, DPMO must
worsen, and either rework or customer returns absorb the difference. A camera system removes
the human throughput ceiling and, by catching faults early in the weaving-to-gluing chain,
prevents value from being added to material that will ultimately be rejected.

\section{Imaging and Acquisition Subsystem}
\label{sec:imaging}
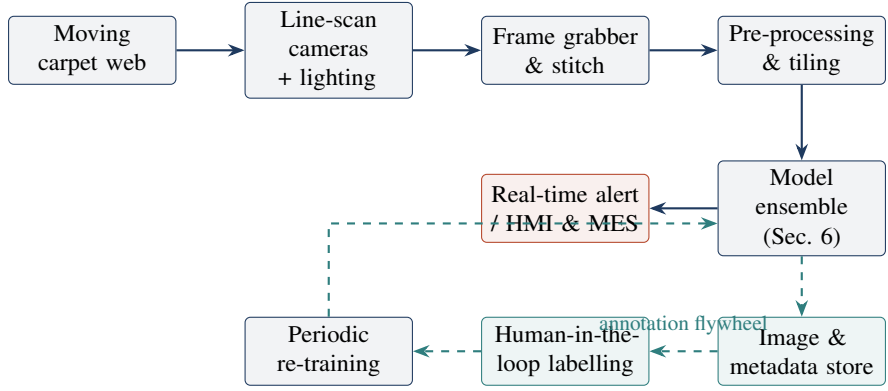
\begin{figure}[t]
\centering
\begin{tikzpicture}[
  font=\footnotesize,
  node distance=6mm and 9mm,
  box/.style={draw=navy,fill=navy!6,rounded corners=2pt,align=center,
              minimum height=9mm,inner sep=3pt,text width=20mm},
  data/.style={draw=teal,fill=teal!8,rounded corners=2pt,align=center,
              minimum height=9mm,inner sep=3pt,text width=20mm},
  act/.style={draw=rust,fill=rust!8,rounded corners=2pt,align=center,
              minimum height=9mm,inner sep=3pt,text width=20mm},
  arr/.style={-{Stealth[length=2.2mm]},navy,thick},
  arr2/.style={-{Stealth[length=2.2mm]},teal,thick,dashed},
]
\node[box] (web) {Moving carpet web};
\node[box,right=of web] (cam) {Line-scan cameras + lighting};
\node[box,right=of cam] (grab) {Frame grabber \& stitch};
\node[box,right=of grab] (pre) {Pre-processing \& tiling};
\node[box,below=10mm of pre] (model) {Model ensemble (Sec.~6)};
\node[act,left=of model] (act) {Real-time alert / HMI \& MES};
\node[data,below=8mm of model] (store) {Image \& metadata store};
\node[data,left=of store] (label) {Human-in-the-loop labelling};
\node[box,left=of label] (train) {Periodic re-training};

\draw[arr] (web)--(cam);
\draw[arr] (cam)--(grab);
\draw[arr] (grab)--(pre);
\draw[arr] (pre)--(model);
\draw[arr] (model)--(act);
\draw[arr2] (model)--(store);
\draw[arr2] (store)--(label);
\draw[arr2] (label)--(train);
\draw[arr2] (train.north) |- ([yshift=-2mm]model.west);
\node[teal,font=\scriptsize] at ($(store)!0.5!(label)+(0,3.5mm)$) {annotation flywheel};
\end{tikzpicture}
\caption{Proposed architecture. The solid (navy/rust) path is real-time inspection; the dashed
(teal) path is the data-collection flywheel that grows a labelled defect dataset and feeds
periodic re-training.}
\label{fig:arch}
\end{figure}

The first engineering decision is how to image a wide web moving continuously without motion
blur, at a resolution fine enough to reveal the smallest fault of interest. Figure~\ref{fig:arch}
shows the proposed end-to-end architecture, including the feedback path that turns inspection
into data collection.

\subsection{Cameras and illumination}
For a continuously moving web, line-scan cameras are the standard choice: each camera captures
one line of pixels at a time and the web's own motion builds the second image dimension, so
there is no motion blur provided the line rate is synchronized to web speed through a shaft
encoder. A single line-scan sensor (typically $8{,}192$ pixels wide, i.e.\ ``8k'') rarely
spans a multi-metre web at fine resolution, so several cameras are mounted side by side and
their fields of view are stitched in software (Figure~\ref{fig:arch}).

Illumination is as important as the sensor. Carpet faults fall into two broad visibility
regimes: \emph{tonal} faults (shade variation, stains, print errors) that are best revealed by
diffuse \emph{bright-field} light, and \emph{structural/topographic} faults (broken picks,
floats, pile crush, holes) that are best revealed by low-angle \emph{grazing} (dark-field)
light that casts shadows from surface relief. We therefore propose a dual-illumination head
that captures the web under both conditions, giving the models complementary channels rather
than forcing a single compromise exposure.

\subsection{Resolution and throughput budget}
\begin{figure}[t]
\centering
\includegraphics[width=0.72\linewidth]{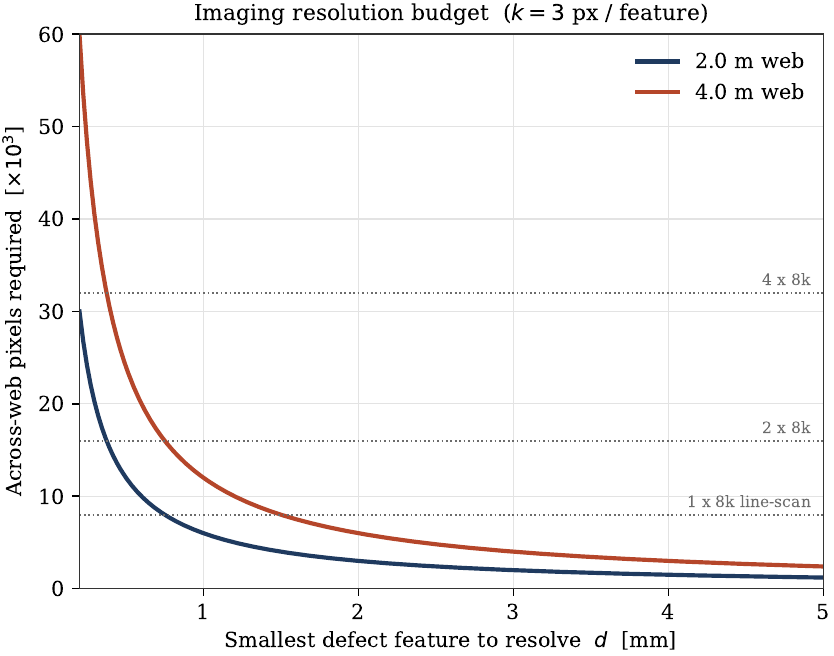}
\caption{Across-web pixel count required to resolve a defect feature of size $d$
(Equation~\eqref{eq:px}, $k=3$). Dotted lines mark the capacity of one, two, and four stitched
8k line-scan cameras. Finer faults and wider webs demand more cameras.}
\label{fig:resbudget}
\end{figure}

Let the web width be $W$ (mm), the smallest defect feature to be resolved be $d$ (mm), and let
$k$ be the number of pixels required to register that feature reliably (a common rule of thumb
is $k \approx 3$). The number of pixels required across the web is
\begin{equation}
N_{\mathrm{px}} \;=\; \frac{k\,W}{d}.
\label{eq:px}
\end{equation}
For $W = 4000$~mm, $d = 0.5$~mm and $k = 3$, Equation~\eqref{eq:px} gives
$N_{\mathrm{px}} = 24{,}000$ pixels across the web, which is met by three stitched 8k cameras.
Figure~\ref{fig:resbudget} plots this relationship for two web widths and makes the camera
count explicit.

The temporal constraint is the line rate. With pixel size $p = d/k$ along the travel direction
and web speed $v$ (mm/s), the required line rate is
\begin{equation}
f_{\mathrm{line}} \;=\; \frac{v}{p} \;=\; \frac{k\,v}{d}.
\label{eq:linerate}
\end{equation}
At $v = 1000$~mm/s ($1$~m/s), $d = 0.5$~mm and $k = 3$, Equation~\eqref{eq:linerate} gives
$f_{\mathrm{line}} = 6000$~lines/s ($6$~kHz), comfortably within the range of industrial
line-scan cameras. The aggregate raw data rate is
$N_{\mathrm{px}} \times f_{\mathrm{line}}$ pixels/s---here $1.44\times10^{8}$ px/s per
illumination channel---which sets the requirements on the frame grabber, on-edge buffering,
and the tiling stage that cuts the continuous web into fixed-size patches for the models.
Table~\ref{tab:imaging} collects representative parameters.

\begin{table}[t]
\centering
\caption{Representative imaging parameters for a wide-web carpet inspection line.}
\label{tab:imaging}
\small
\begin{tabular}{@{}l l l@{}}
\toprule
\textbf{Parameter} & \textbf{Symbol} & \textbf{Representative value}\\
\midrule
Web width & $W$ & $4000$~mm\\
Target feature size & $d$ & $0.5$~mm\\
Pixels per feature & $k$ & $3$\\
Across-web resolution & $N_{\mathrm{px}}$ & $24{,}000$ px ($3\times$ 8k)\\
Web speed & $v$ & $1.0$~m/s\\
Required line rate & $f_{\mathrm{line}}$ & $6$~kHz\\
Illumination & --- & bright-field $+$ grazing\\
Raw rate / channel & --- & $\approx 1.4\times10^{8}$ px/s\\
\bottomrule
\end{tabular}
\end{table}

\section{A Carpet Defect Taxonomy}
\label{sec:taxonomy}
A useful dataset starts from a clear, shared vocabulary of faults. Table~\ref{tab:taxonomy}
proposes a taxonomy organized by origin in the weaving-to-gluing process, with the
illumination channel most likely to reveal each family and the model type best suited to it.
The taxonomy is deliberately practical: each class corresponds to a label an inspector can
apply and a model can learn, and it spans the spectrum from sharp, localized faults (well
suited to bounding-box detection) to diffuse, area-extended faults (well suited to
segmentation).

\begin{table}[t]
\centering
\caption{Proposed carpet defect taxonomy. Channel: B = bright-field, G = grazing.
Model: D = detection, S = segmentation, A = anomaly score.}
\label{tab:taxonomy}
\small
\begin{tabular}{@{}l l l c c@{}}
\toprule
\textbf{Family} & \textbf{Examples} & \textbf{Typical cause} & \textbf{Channel} & \textbf{Model}\\
\midrule
Structural & broken/missing pick, float, double pick & loom / yarn break & G & D\\
Pile & crush, matting, bald spot, high pile & wear, tension & G & D/S\\
Colour/dye & shade band, streak, off-shade & dye lot, yarn mix & B & S\\
Contamination & oil mark, stain, foreign fibre & handling, debris & B & D\\
Holes/tears & hole, cut, snag & mechanical damage & B/G & D\\
Pattern & misalignment, skew, registration & setup, distortion & B & S\\
Backing & delamination, adhesive starvation, bubble & gluing stage & G & S\\
\bottomrule
\end{tabular}
\end{table}

\section{Dataset Construction and the Annotation Flywheel}
\label{sec:data}
\begin{figure}[t]
\centering
\includegraphics[width=0.72\linewidth]{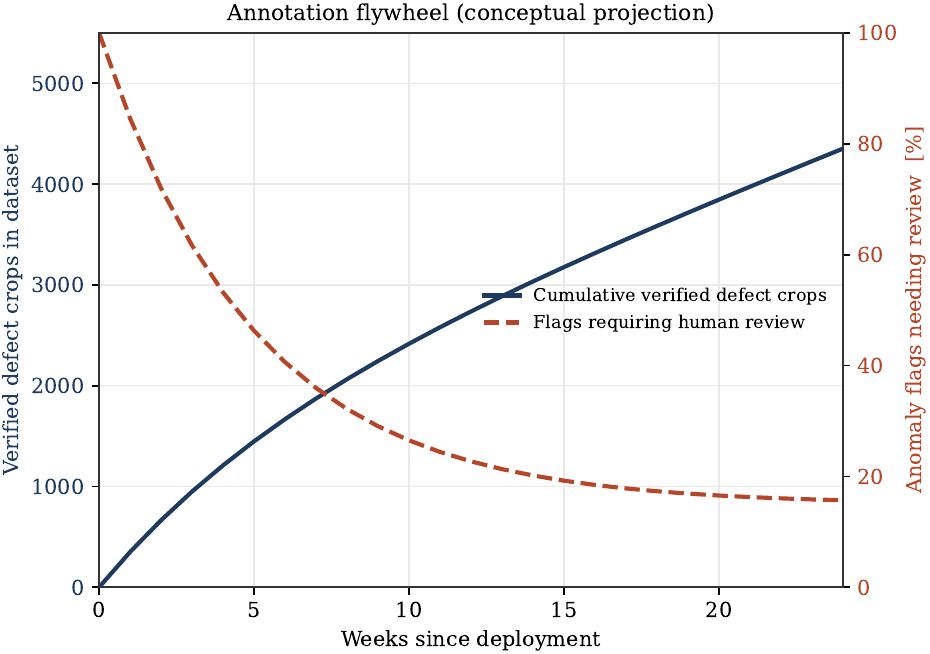}
\caption{Conceptual projection of the annotation flywheel. As verified defect crops accumulate
(left axis), the share of anomaly flags that still require human adjudication falls (right
axis) because the maturing supervised models resolve routine cases automatically. The curves
are illustrative of the intended dynamics, not measured results.}
\label{fig:flywheel}
\end{figure}

The defining difficulty of industrial defect data is asymmetry: defect-free material is
effectively unlimited, while each specific defect is rare, and some classes may appear only a
handful of times per month. A naive plan---``collect a large labelled dataset, then train a
model''---fails because the labelled defects do not yet exist and would take many months of
manual annotation to accumulate. We instead propose a staged, self-reinforcing process,
illustrated conceptually in Figure~\ref{fig:flywheel}.

\paragraph{Stage 0 --- Normal-only bootstrap.}
From the first day of running, the system records defect-free web under both illumination
channels. This requires no labels and immediately yields the large ``normal'' corpus needed by
the anomaly-detection models of Section~\ref{sec:models}. The carpet category of MVTec~AD shows
that this regime is both realistic and benchmarked~\cite{mvtec2019}.

\paragraph{Stage 1 --- Anomaly-driven candidate mining.}
An unsupervised detector (Section~\ref{sec:models}) scores every tile. High-scoring tiles are
saved as \emph{candidates} with their score, location, time, loom ID, and illumination
channel. Most candidates are real faults or interesting borderline cases; this concentrates
human attention on the few percent of material worth labelling rather than on the whole web.

\paragraph{Stage 2 --- Human-in-the-loop labelling.}
A lightweight review interface presents candidate crops to an inspector, who confirms or
rejects the fault and assigns a class from Table~\ref{tab:taxonomy} (and, where useful, a
polygon for segmentation). Verified crops accumulate into a growing supervised dataset.
Because candidates are pre-filtered, the labelling cost per verified defect is far lower than
exhaustive frame-by-frame annotation.

\paragraph{Stage 3 --- Augmentation and class balancing.}
Rare classes are amplified with geometric and photometric augmentation and, importantly, with
\emph{defect synthesis}: real defect crops are composited onto defect-free backgrounds, and
self-supervised schemes such as CutPaste~\cite{cutpaste} and DRAEM~\cite{draem} generate
additional plausible anomalies. This counteracts class imbalance without waiting for rare
faults to recur naturally.

\paragraph{Stage 4 --- Periodic re-training and active learning.}
At intervals, the supervised models are re-trained on the enlarged dataset, and the anomaly
detector's memory bank is refreshed with current normal material to track drift in yarn,
colour, and pattern. The system preferentially surfaces uncertain or novel cases (active
learning), so each labelling hour buys the maximum reduction in model error.

This design makes data collection the system's primary long-term product. The cameras are not
merely a sensor for a fixed model; they are the instrument that builds an ever-better,
plant-specific dataset---one tuned to the plant's own yarns, dyes, looms, and patterns rather than
to a generic public corpus.

\section{Modelling Strategy for Quality Control}
\label{sec:models}
We propose an ensemble whose composition shifts as the dataset matures
(Table~\ref{tab:models}). Each model family answers a different question and operates on the
fixed-size tiles produced by the pre-processing stage.

\paragraph{Unsupervised anomaly detection (available first).}
Trained only on defect-free tiles, methods such as PaDiM~\cite{padim} and
PatchCore~\cite{patchcore} produce a per-tile anomaly score and a coarse heat map by comparing
deep features against a model of normality; convolutional autoencoders provide a
reconstruction-error alternative~\cite{mei2018,mvtec2019}. This family carries the system from
day one, before any defect labels exist, and continues to act as a safety net for novel,
never-before-seen faults.

\paragraph{Supervised classification (early).}
Once a few hundred verified crops per class exist, a backbone such as
ResNet~\cite{resnet} or EfficientNet~\cite{efficientnet}, fine-tuned by transfer learning,
classifies a flagged tile into a defect family from Table~\ref{tab:taxonomy}. Classification is
cheap to train and provides the family labels that drive the DMAIC ``Analyse'' phase.

\paragraph{Supervised detection (mid).}
With enough localized labels, a detector from the YOLO family~\cite{yolo,electronics2025} or
Faster~R-CNN~\cite{fasterrcnn} returns bounding boxes for sharp, localized faults (broken
picks, holes, contamination), enabling precise flagging and counting.

\paragraph{Supervised segmentation (mature).}
For area-extended faults (shade bands, delamination, pattern skew), pixel-level models such as
U-Net~\cite{unet} or Mask~R-CNN~\cite{maskrcnn} delineate the affected region, which is what
downstream grading and trim-decision logic actually needs.

\begin{table}[t]
\centering
\caption{Model families, their supervision, output, and the dataset maturity at which they
become useful.}
\label{tab:models}
\small
\begin{tabular}{@{}l l l l@{}}
\toprule
\textbf{Family} & \textbf{Supervision} & \textbf{Output} & \textbf{Available}\\
\midrule
Anomaly detection & normal-only & score + heat map & day one\\
Classification & labelled crops & defect family & early\\
Detection & boxes & located faults & mid\\
Segmentation & masks & pixel fault map & mature\\
\bottomrule
\end{tabular}
\end{table}

Operationally, the anomaly detector runs always and gates the supervised models; a tile that
the supervised models classify with low confidence but the anomaly detector scores highly is
routed to human review, which both protects quality and feeds Stage~2 of the flywheel.

\section{Linking Detection Performance to Six Sigma Outcomes}
\label{sec:sixsigma}
The business value of the system is best expressed in the language of the charter. The process
sigma level is a standardized measure of capability derived from the defect rate, with DPMO
defined as
\begin{equation}
\mathrm{DPMO} \;=\; \frac{\text{number of defects}}{\text{units}\times\text{opportunities per unit}}\times 10^{6}.
\label{eq:dpmo}
\end{equation}
Higher sigma corresponds to lower DPMO; the canonical Six Sigma target of $3.4$~DPMO is the
aspirational endpoint of the scale~\cite{pyzdek}. The charter recorded a baseline of about
$2.0\times10^{4}$~DPMO and a reported sigma level of $2.33$.

The relevant point is directional and robust: \emph{escaped} defects---faults that reach the
customer or the next stage undetected---are what Equation~\eqref{eq:dpmo} counts. Manual
inspection at high line speed inspects only a fraction of the surface, so its escape rate rises
as throughput rises. Full-coverage camera inspection breaks that coupling. If automated
inspection catches a fraction $r$ of the faults that previously escaped, the escaped-defect
count---and hence DPMO---falls roughly in proportion to $(1-r)$, moving the process up the
sigma scale. Because the system also catches faults early (at weaving rather than after
gluing), it additionally reduces the \emph{cost} of each caught defect, since less value has
been added to condemned material. Set against the charter's $\approx 50{,}000$~EUR/day
exposure and the $40\%$ planned throughput increase, even a moderate improvement in capture
rate is economically decisive, while the system removes the human inspection ceiling that the
new looms would otherwise overwhelm.

\section{Deployment Considerations}
\label{sec:deploy}
\paragraph{Edge inference and latency.} To act in real time, inference should run on an
industrial GPU or edge accelerator beside the line; recent work demonstrates real-time textile
inspection on embedded hardware~\cite{electronics2025}. The anomaly detector and a compact
detector run synchronously with the web; heavier segmentation can run asynchronously on flagged
regions.

\paragraph{Integration and human roles.} Defect events, classes, locations, and confidences
should be logged to the plant's manufacturing execution system (MES) for traceability and
statistical process control, sustaining the DMAIC ``Control'' phase. Inspectors are not removed
but redeployed: from scanning every metre of web to adjudicating flagged cases and labelling,
which is exactly the Stage~2 activity that improves the models.

\paragraph{Drift and maintenance.} Yarn lots, dye batches, and patterns change, so ``normal''
drifts. The normal corpus and the anomaly detector's reference set must be refreshed on a
schedule, and model performance must itself be monitored---a model whose flag rate suddenly
changes is a signal worth investigating.

\section{Limitations and Future Work}
\label{sec:limits}
This is a design proposal, not an evaluated deployment; the figures are engineering budgets and
conceptual projections rather than measured results, and real capture rates depend on the
plant's specific products and lighting. Several risks deserve attention. Highly patterned or
multi-colour carpet complicates the notion of ``normal'' and may require per-pattern reference
models. Very rare or visually subtle faults may resist even the anomaly detector and will
depend heavily on defect synthesis and active learning. Stitching multiple cameras and
balancing dual illumination introduce calibration and photometric-consistency work that should
not be underestimated. Future work includes a pilot on a single loom to measure capture rate
and false-alarm rate against an inspector baseline, an ablation over illumination channels, and
a cost model that converts measured capture rate into a DPMO and sigma-level estimate specific
to the deployment site.

\section{Conclusion}
\label{sec:conclusion}
We have set out an end-to-end proposal for camera-based, AI-assisted quality control in carpet
manufacturing that treats defect-pattern data collection as a first-class engineering goal. The
proposal specifies a dual-illumination, multi-camera line-scan imaging system with an explicit
resolution and throughput budget; a practical carpet defect taxonomy; a staged modelling
strategy that begins with unsupervised anomaly detection and matures into supervised detection
and segmentation; and an annotation flywheel that turns ordinary production into an
ever-improving, plant-specific dataset. Grounded in a documented Six Sigma project in woven
carpet production, the
approach maps directly onto the DMAIC phases and onto the charter's DPMO, throughput, and
financial objectives: by replacing sampled manual inspection with full-coverage automated
inspection, it removes the inspection bottleneck the new looms would create, lowers escaped
defects and therefore DPMO, and builds the data asset that makes each successive model better
than the last.

\section*{Acknowledgements}
The author thanks the SAG team for the samples and the environment in which to write this
paper.

\end{document}